\documentclass{article} % For LaTeX2e
\usepackage{iclr2025_conference,times}

\usepackage{amsmath,amsfonts,bm}

\def\eqref#1{equation~\ref{#1}}
\def\1{\bm{1}}

\DeclareMathAlphabet{\mathsfit}{\encodingdefault}{\sfdefault}{m}{sl}
\SetMathAlphabet{\mathsfit}{bold}{\encodingdefault}{\sfdefault}{bx}{n}

\usepackage{import}
\usepackage{hyperref}
\usepackage{url}
\usepackage{xcolor}
\usepackage[colorinlistoftodos,prependcaption]{todonotes}
\usepackage{xargs} % Required for the \newcommandx
\usepackage{booktabs}
\usepackage{titling}
\usepackage{float} 

\title{Prithvi WxC: Foundation Model for Weather and Climate
}

\author{%
    \begin{minipage}{\textwidth} 
    \centering
    \textbf{Johannes Schmude}\textsuperscript{1,$\dagger, \ddagger$}, \textbf{Sujit Roy}\textsuperscript{2,7,$\dagger, \ddagger$}, \textbf{Will Trojak}\textsuperscript{1}, \textbf{Johannes Jakubik}\textsuperscript{1}, \textbf{Daniel Salles Civitarese}\textsuperscript{1}, \textbf{Shraddha Singh}\textsuperscript{1}, \textbf{Julian Kuehnert}\textsuperscript{1}, \textbf{Kumar Ankur}\textsuperscript{2}, \textbf{Aman Gupta}\textsuperscript{3}, \textbf{Christopher E Phillips}\textsuperscript{2}, \textbf{Romeo Kienzler}\textsuperscript{1}, \textbf{Daniela Szwarcman}\textsuperscript{1},  \textbf{Vishal Gaur}\textsuperscript{2}, \textbf{Rajat Shinde}\textsuperscript{2}, \textbf{Rohit Lal}\textsuperscript{2}, \textbf{Arlindo Da Silva}\textsuperscript{6}, \textbf{Jorge Luis Guevara Diaz}\textsuperscript{1}, \textbf{Anne Jones}\textsuperscript{1}, \textbf{Simon Pfreundschuh}\textsuperscript{4}, \textbf{Amy Lin}\textsuperscript{2}, \textbf{Aditi Sheshadri}\textsuperscript{3}, \textbf{Udaysankar Nair}\textsuperscript{2}, \textbf{Valentine Anantharaj}\textsuperscript{5}, \textbf{Hendrik Hamann}\textsuperscript{1}, \textbf{Campbell Watson}\textsuperscript{1}, \textbf{Manil Maskey}\textsuperscript{7}, \textbf{Tsengdar J Lee}\textsuperscript{8}, \textbf{Juan Bernabe Moreno}\textsuperscript{1}, \textbf{Rahul Ramachandran}\textsuperscript{7} \\
     \vspace{0.5cm} 
    \textsuperscript{$\dagger$}Equal Contribution, \\\textsuperscript{$\ddagger$}Johannes.Schmude@ibm.com , Sujit.Roy@nasa.gov
     \thanks{\textsuperscript{1}IBM Research; \textsuperscript{2}Earth System Science Center,
The University of Alabama in Huntsville, Huntsville, AL, USA; 
\textsuperscript{3}Department of Earth System Science, Stanford University, Stanford, USA;
\textsuperscript{4}Department of Atmospheric Science, Colorado State University, Fort Collins, CO, USA;
\textsuperscript{5} National Center for Computational Sciences / Oak Ridge National Laboratory;
\textsuperscript{6}NASA/Goddard Space Flight Center, USA;
\textsuperscript{7}NASA Marshall Space Flight Center, Huntsville, AL, USA;
\textsuperscript{8}NASA Headquarters, Washington, DC, United States of America
    }
    \end{minipage}
}
\date{} 

\iclrfinalcopy % Uncomment for camera-ready version, but NOT for submission.
\begin{document}

\maketitle

\begin{abstract}

Triggered by the realization that AI emulators can rival the performance of traditional numerical weather prediction models running on HPC systems, there is now an increasing number of large AI models that address use cases such as forecasting, downscaling, or nowcasting. While the parallel developments in the AI literature focus on foundation models -- models that can be effectively tuned to address multiple, different use cases -- the developments on the weather and climate side largely focus on single-use cases with particular emphasis on mid-range forecasting. We close this gap by introducing Prithvi WxC, a 2.3 billion parameter foundation model developed using 160 variables from the Modern-Era Retrospective Analysis for Research and Applications, Version 2 (MERRA-2). Prithvi WxC employs an encoder-decoder-based architecture, incorporating concepts from various recent transformer models to effectively capture both regional and global dependencies in the input data. The model has been designed to accommodate large token counts to model weather phenomena in different topologies at fine resolutions. Furthermore, it is trained with a mixed objective that combines the paradigms of masked reconstruction with forecasting. We test the model on a set of challenging downstream tasks namely: Autoregressive rollout forecasting, Downscaling, Gravity wave flux parameterization, and Extreme events estimation. The pretrained model with 2.3 billion parameters, along with the associated fine-tuning workflows, has been publicly released as an open-source contribution via Hugging Face.
\end{abstract}

\section{Introduction}

Deep learning is increasingly transforming weather applications by delivering highly accurate forecasts with reduced computational costs compared to traditional numerical weather prediction methods \citep{bi2023accurate,lam2023learning, mukkavilli2023ai}. Unlike the traditional physics-based approaches, deep learning models do not directly simulate the underlying physics. Instead, they capture this through probability distributions derived from model training, a method adapted from natural language processing and computer vision. This technique has proven surprisingly effective in approximating complex physical systems such as the weather. However, most current deep learning models for weather are \textit{task-specific} forecast emulators, which focus solely on the forecasting problem. (See, however, \cite{koldunov2024emerging}.) Key examples include FourCastNet \citep{pathak2022fourcastnet}, Pangu \citep{bi2022pangu}, GraphCast \citep{lam2022graphcast}, FengWu \citep{chen2023fengwu}, Stormer \citep{nguyen2023scaling} and AIFS \citep{lang2024aifs}. Machine learning models also show promise for longer-term subseasonal-to-seasonal forecasts \citep{Weyn.etal2021}. Additionally, ML-based approaches are being explored to enhance climate predictions \citep[see][for a review]{Mansfield.etal2023, Eyring.etal2024}, with a focus on the development of ML-driven parameterizations \citep{Rasp.etal2018, Zhao.etal2019,Espinosa.etal2022,Yuval.O’Gorman2023, Henn.etal2024, gupta2024machine}, bias corrections \citep{Bretherton.etal2022, Gregory.etal2024}, and assessments of climate change impacts \citep[among others]{Davenport.Diffenbaugh2021, Diffenbaugh.Barnes2023}. There is fascinating emerging work that combines the strengths of the data-driven and physics-based approaches \citep{kochkov2024neural,husain2024leveraging, roy2024clifford}. Finally, there are further large, task-specific models for Nowcasting \citep{andrychowicz2023deep} and downscaling \citep{mardani2024residual}.

Looking beyond atmospheric sciences at developments in AI in general and language models in particular, the last few years have been dominated by the emergence of foundation models. That is, large AI models pretrained in a task-agnostic manner that can be effectively fine-tuned to address a number of specific use cases. Despite the mirroring successes of large AI models in both fields, applications of the foundation model principle to atmospheric sciences have been rare. AtmoRep \citep{lessig2023atmorep} considered problems ranging from nowcasting to downscaling and bias corrections; Aurora \citep{bodnar2024aurora} focusses a number of different \emph{forecasting} problems.

% Intro model + data
To address this gap, we introduce Prithvi WxC, a large-scale foundation model for weather and climate applications trained on 160 atmospheric variables from the Modern-Era Retrospective analysis for Research and Applications, Version 2 (MERRA-2) data set. MERRA-2 is a widely-used reanalysis dataset from NASA providing global atmospheric data, including temperature, humidity, and wind. Spanning from 1980 to the present day with spatial resolution of 0.5 by 0.625 degrees and temporal resolution of 3 hours \citep{gelaro2017modern}, it is valuable for climate research and atmospheric studies.

% Architecture
Prithvi WxC is a transformer-based deep learning architecture which combines ideas from several recent transformer architectures in order to effectively process regional and global dependencies of the input data and to efficiently process longer sequence lengths of tokens. This allows the model to, for example, run in different spatial contexts or infuse additional tokens from off-grid measurements to the model during finetuning.
We additionally experiment with different loss functions, for example, by removing task-specific temporal variances from loss functions of forecast emulators and replacing them with task-agnostic climatology variances.

% $V_C$. With this change in the optimization target, we prevent the target to collapse for non-forecasting tasks, where temporal variances are not applicable. We introduce our climatology-based optimisation target in equation~(1), where $f_{\theta}\left(\frac{X(t)-\mu}{\sigma} ; C(t+\delta t)  ; \delta t\right)$ denotes the output of our model given normalised inputs and climatology data. The climatology variances $V_C$ scale several factors within the loss function and, thereby, serve as a prior to the model.

% \begin{equation}
% V_C^{-1} \times\left\{X(t+\delta t)-\left[\sqrt{V_C} \times f_\theta\left(\frac{X(t)-\mu}{\sigma} ; C(t+\delta t) ; \delta t\right)+C(t+\delta t)\right]\right\}
% \end{equation}

% Scaling
% Further, we are experimenting with scaling Prithvi WxC to larger parameter counts in order to better understand whether the capabilities of the model change with an increasing number of weights. For that we shard the data via the fully-sharded data parallel (FSDP) framework and train the model across dozens of GPUs for multiple days on NASA Advanced Supercomputing (NAS) clusters. By experimenting with how far we can effectively scale the model, we naturally address the inherent tradeoff between a higher number of model parameters and larger batch sizes during pretraining.  

% Validation
The validation of Prithvi WxC extends from zero shot evaluations for reconstruction and forecasting to downstream tasks such as downscaling of weather and climate models, the prediction of hurricane tracks and atmospheric gravity wave flux parameterization.

\section{Prithvi WxC}

From an AI perspective, Prithvi WxC has been designed to address several questions that arise when considering the meaning of foundation models for atmospheric physics: Since weather models can run on the entire earth or in a regional context, do we need specialized architectures for global and local problems? Do we need to differentiate between models with zero and non-zero lead time? If we do consider tasks with zero and non-zero lead time, what is a suitable pretext task for pretraining?

\subsection{Pretraining objective}

As outlined above, the most celebrated successes at the intersection of AI and atmospheric sciences concern \emph{forecast emulators}. I.e.~models that are given the state of the atmosphere at times $t$ and $t-\delta t$ predict the state at $t+\delta t$. While forecasting is a obvious task for weather and climate data, the prototypical approach in the computer vision literature is that of the masked autoencoder (MAE) \citep{he2022masked}. There are several considerations that make masking attractive for pretraining:

To start, while both NWP as well as reanalysis data is gridded and dense, observational data is ungridded and sparse. As such, it might not be surprising that the emerging literature on models working directly on observation makes heavy use of masking \citep{vandal2024global, mcnally2024data}. On a related note, the forecasting objective becomes trivial in the case of $\delta t = 0$. At the same time there are use cases such as downscaling or data assimilation for which such a time step is meaningful. Thus, a \emph{foundation model} aiming to address all such use cases should be able to deal with a non-positive forecast step. On the more technical side, a common problem in this space is the size of the data. As noted in the original work \citep{he2022masked}, masking is highly memory efficient. As long as masking is implemented without additional masking tokens, the technique reduces memory pressure and increases training speeds.

To our knowledge, only \citep{lessig2023atmorep, schmude2023efficient} used masking on reanalysis data. The latter did so in the context of contrastive learning while the former used a 3D masking approach akin to \citep{feichtenhofer2022masked}. However, given the success of the forecasting objective and in order to avoid holding a 3D data cube in GPU memory, we merge these objectives slightly differently. Effectively we train a 2D model with masking for which the output is a prediction:
\begin{equation}
    \label{eq:pretraining_objective_simplified}
    \hat{X}_{t+\delta t} = f_\theta \left\lbrack M_{0.5} \left( X_t, X_{t-\delta t} \right) \right\rbrack.
\end{equation}
Here $X_t$ is the data, $\hat{X}_t$ a prediction, $f_\theta$ a neural network and $M_{0.5}$ the masking operator for $50\%$.

Now, given that we are training a foundation model, we are aiming for easy generalizability. Thus, we allow for different values for $\delta t$ for inputs and the target. A variable lead time appeard already in \cite{nguyen2023scaling}. Moreover, we also use some static data such as elevation, land fraction etc. See \ref{sec:dataset} for details.

The objective of \eqref{eq:pretraining_objective_simplified} can be improved on. Indeed, the forecast emulators of \citep{lam2023learning,nguyen2023scaling} do not predict $X_{t+\delta t}$ but the tendency $X_{t+\delta t} - X_{t}$. (See however \citep{lang2024aifs} as well as \citep{bodnar2024aurora} which both output absolute quantities.) The rationale seems clear: A model that predicts tendencies gets the performance of a persistence forecast for free and one does not spend a lot of GPU cycles learning known biases. In our case however the objective would again become trivial for $\delta t = 0$.

In light of this we turn to another source of free information: Climatology. That is, instead of predicting the difference from the current time stamp, we model the deviation from historical climate at this time, $C_t$. All in all, our pretraining objective is
\begin{equation}
    \label{eq:pretraining_objective}
    \frac{\hat{X}_{t+\delta t} - C_{t+\delta t}}{\sigma_C} = f_\theta\left\lbrack M_{0.5}\left( \frac{X_t-\mu}{\sigma}, \frac{X_{t-\delta \tau}-\mu}{\sigma} \right); \frac{C_{t+\delta t}-\mu}{\sigma}, S, \delta t, \delta \tau \right\rbrack.
\end{equation}
Here, $\mu$ and $\sigma$ are per parameter means and standard deviations (computed across space and time). $\sigma_C^2 = \sigma_C^2 (X_t - C_t)$ is the variance of the historical anomaly; again comptued aross space and time. $S$ are static inputs and $\delta t$ and $\delta \tau$ are the time steps for the target and the inputs respectively.

\subsection{Data}
\label{sec:dataset}

\subsubsection{MERRA-2}

The Modern-Era Retrospective Analysis for Research and Applications Version 2 (MERRA-2) \citep{gelaro2017modern}, developed by NASA's Global Modeling and Assimilation Office (GMAO), serves as the primary dataset for this study. It uses a cubed-sphere grid, which results in uniform grid spacing at all latitudes. This design minimizes grid spacing irregularities found in latitude-longitude grids, enhancing the dataset's spatial consistency and usefulness for global-scale analyses. MERRA-2 provides a comprehensive and consistent record of Earth's climate and atmospheric conditions, offering valuable insights into long-term climate trends and variability. It is a state-of-the-art reanalysis dataset that integrates a range of observational data with advanced modeling techniques to produce a high-quality, multidecadal record of atmospheric conditions \citep{rienecker2011merra, gelaro2017modern}. It is particularly useful for climate research due to its extensive historical coverage and sophisticated data assimilation methods.

The dataset includes variables at model native levels corresponding to nominal pressure surfaces which are 985 hPa, 970 hPa, 925 hPa, 850 hPa, 700 hPa, 600 hPa, 525 hPa, 412 hPa, 288 hPa, 245 hPa, 208 hPa, 150 hPa, 109 hPa, and 48 hPa, with data available every 3 hours. Variables at these levels include wind components (U, V), vertical wind ($\omega$), air temperature (T), specific humidity (QV), actual mid-level pressure (PL), and mid-layer geopotential height (H), cloud fraction (CLOUD), cloud massk fraction that is ice (QI) and water (QL). 

Additional single-level variables are available at 1-hour intervals and include near-surface wind components (U10, V10), near-surface (2 meter) air temperature (T2M), skin temperature (TS), surface roughness (Z0M), specific humidity (QV2M), surface pressure (PS), sea level pressure (SLP), column-total ice, liquid water and water vapor (TQI, TQL, TQV), longwave radiation emitted by the surface (LWGEM), longwave radiation absorbed by the surface (LWGAB), upward longwave at the top of atmosphere (LWTUP), net downward shortwave radiation at the surface (SWGNT) and net shortwave at top of atmosphere (SWTNT). Static variables include surface geopotential height (PHIS), land fraction (FRLAND), ocean fraction (FROCEAN), and ice fraction (FRACI), which are used to provide essential static information, and is varying in space, but not time. Time-averaged variables, such as rootzone soil wetness (GWETROOT), leaf area index (LAI), and surface fluxes (EFLUX, HFLUX), are aggregated from 1-hourly intervals, because these are the diagnostics variables and not available at the analysis time. Aggregation methods are used for variables from hourly products, where means of adjacent hourly values are used to create 12:00 UTC data. For example, the mean of 11:30 and 12:30 values is calculated to prepare the 12:00 UTC data. Missing values (NaNs) in GWETROOT and LAI are replaced with 1 and 0, respectively, to maintain data availability over the ocean. Static datasets are incorporated by creating monthly files, ensuring that the static variables (PHIS, FRLAND, FROCEAN, FRACI) remain consistent for each month, thereby maintaining the integrity of static information throughout the dataset. List of variables used in the training is listed in the Appendix tables \ref{tab:variables}, \ref{tab:variables2} and \ref{tab:variables3}. We train the model using data from 1980 to 2019. We validate with data from one of the years in the 2020-2023 range, depending on task.

\subsubsection{Climatology}

The climatology appearing in \eqref{eq:pretraining_objective} was computed from 20 years of MERRA-2 data following the methodology of the ERA-Interim climatology \citep{janousek2011eraintclimatology}. That is, for each Julian day and each hour of the day we aggregate all data across the last 20 years. Subsequently we apply a 61-day rolling window weighted average to this. The weights are given by a second order polynomial. Thus the climatology resolves the day-night cycle. There are $365 \times 8$ timestamps and each pixel is based on $20 \times 61 = 1220$ data points. We used the same 20 year period that we used for training; that is 1980-2019.

\subsubsection{Normalization}

While \eqref{eq:pretraining_objective} is a fairly natural training objective, we found that leaving the normalization constants $\sigma$ and $\sigma_C$ unconstrained leads to instabilities during training. This is essentially due to the large range of values we have, especially the anomalies in the mass fraction of cloud liquid water QL at high model levels can be as small as $10^{-26}$ at level 34. To avoid such extreme values upsetting numerics, we impose $10^{-4} \leq \sigma \leq 10^4$ and similarly $10^{-7} \leq \sigma_C \leq 10^7$. In both cases, this mainly affects $Q_I$ and $Q_L$ at high levels.

\subsection{Architecture}

\subsubsection{A scalable and flexible vision transformer}

At it's core, Prithvi WxC is a scalable and flexible 2D vision transformer. To keep it as flexible as possible, we aimed not to use architecture elements that restrict to ``rectangular'' topologies for data. (Even though we train on MERRA-2 data on a rectangular lat/lon grid, one can envision training or running inference directly on Gaussian grids.) Vanilla ViTs would satisfy this requirement, yet do not scale to large token counts. Considering the different flavors of scalable transformers we notice the findings of ``Hiera'' \citep{ryali2023hiera}. Here the authors show that it is possible to surpass the performance of Swin transformers with a more flexible and simplified architecture. Turning back to AI models for weather, \cite{andrychowicz2023deep} made use of MaxViT \citep{tu2022maxvit} which leverages axial attention (albeit with convolutions). In the end, our core idea is that if we pretrain the model using only attention, we keep the core of the model flexible and can add convolutions at fine-tuning time to increase performance when suitable. We do so by joining the approaches of Hiera and MaxViT.

\begin{figure}[h]
    \centering
    \includegraphics[width=1.0\textwidth]{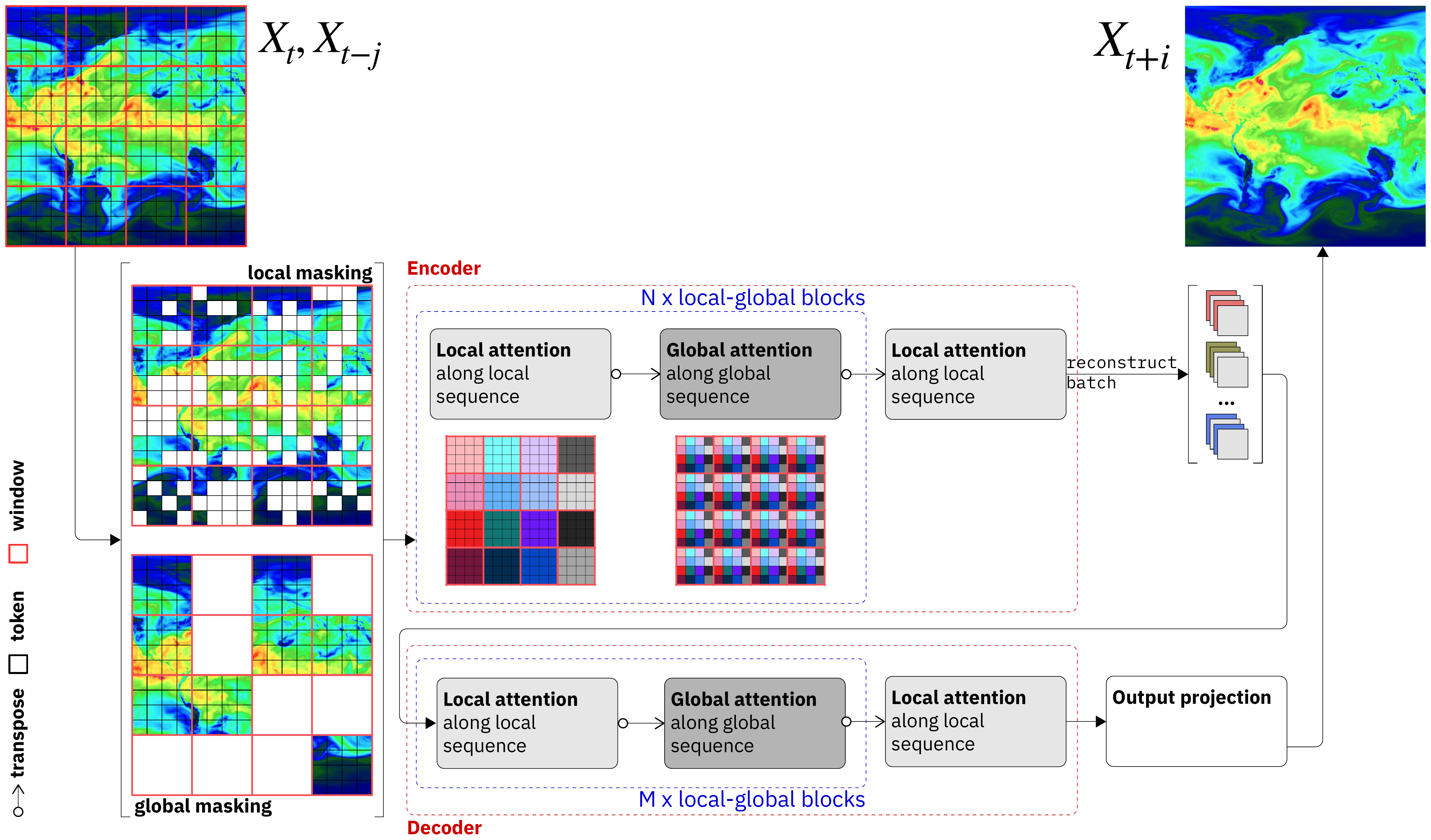}
    \caption{Prithvi WxC core architecture elements and masking scheme. For simplicity the figure ignores elements such as embedding and output layers as well as position encodings.}
    \label{fig:prithvi_wxc_model_architecture}
\end{figure}

In detail, the only constraint we impose on the data is the ability to structure tokens into windows -- akin to Swin, Hiera and MaxViT. Doing so, our data takes the shape (batch, windows, tokens, features), where the second dimension enumerates windows and the third tokens within each window. Subsequently we alternate attention within a window and across windows. The latter is similar to \citep{tu2022maxvit}: Modulo masking the nth token in each window interacts with the nth token in all other windows. This is easily implemented by transposing the window and token dimension between transformer layers. Attention acts on the $3^{rd}$ dimension with the second being rolled into the batch dimension. In what follows we will refer to attention within a window as ``local'' and attention across windows as ``global''. When masking, we can either mask out entire global windows or individual tokens within a window. A byproduct of the latter is that global attention no longer connects the same token in each window. E.g.~if token 1 in window 1 is masked, then token 2 in window 1 connects to token 1 in window to etc. For an illustration of the attention pattern as well as the overall encoder/decoder architecture see figure \ref{fig:prithvi_wxc_model_architecture}.

\subsubsection{Overall model architecture}

As shown in \eqref{eq:pretraining_objective}, the model has several inputs: To start, there are the \emph{dynamic} inputs $X_t$, $X_{t-\delta \tau}$ from the MERRA-2 re-analysis. These take the shape $T \times \lbrack P_S + (P_V \times L) \rbrack \times H \times W$. Here, $T$ is time, fixed to 2. $P_S$ are the 20 surface parameters and $P_V$ the 10 vertical parameters at $L=14$ vertical model levels for a total of 160 inputs. $H$ and $W$ denote latitude and longitude respectively. Since we flatten all temporal and parameter dimensions, the dynamic model input comprises a data cube of dimension $320 \times 360 \times 576$. Climatological inputs $C_{t+\delta\tau}$ take the shape $160 \times 360 \times 576$ as there is the same number of parameters yet no time dimension. The static inputs are based on 4 static parameters from MERRA-2 -- elevation, land cover, ice cover, lake cover -- as well as cosine and sine of day of year and hour of the day.

We use a static Fourier position encoding that respects the periodicity of the earth. In addition, there is a learned encoding for both the lead time $\delta t$ as well as the input time step $\delta \tau$. An earlier version of the model used a non-spatiotemporal context token to communicate information such as the lead time. However, this led to the emergence of specialized transformer layers that paid heavy attention to this token, which is in conflict with stochastic depth (drop path) which we enabled during the scaling phase. Finally there are separate linear embedding layers for the dynamic inputs as well as the concatenation of the climatological and static ones. Once embedded, all tokens are added up.

\subsection{Pretraining}

\subsubsection{Scaling}

In its final configuration, Prithvi WxC comprises 25 encoder and 5 decoder blocks. As both the encoder and decoder start and end with local attention, 13 (3) of these blocks perform local and 12 (2) global attention respectively. The internal dimension is 2,560. With 16 attention heads and an MLP multiplier of 4 this results in 2.3 billion parameters. We use a token size of 2 by 2 pixel. Each window measures 30 by 32 pixel or 15 by 16 tokens. With these choices we are dealing with 51,840 tokens per sample yet are keeping the length of the global and local sequence roughly balanced. Note that both token and window size can be changed when tuning the model and we will do so repeatedly below. In either case, with these choices, the model consumes a bit more than 43 GB of GPU memory in pretraining. If we keep the masking at 50\% we are able to backpropagate through 4 autoregressive steps on a 80 GB A100. (Masking only applies to the first autoregressive step.) If the data becomes dense (i.e.~0\% masking) this reduces to 3 steps. Since the data becomes dense in the decoder and our pretraining data does live on a rectangular grid, we add a Swin-shift to the decoder layers. The overall scale was chosen to ensure that autoregressive ``rollout'' training is still possible.

To bring the model to this scale, we make use of Fully Sharded Data Parallelism (FSDP) as well as flash attention (via scaled dot product attention). We train the model with bfloat16 precision. However, to ensure numeric stability we only use bfloat16 for the transformer layers. The input and output layers remain at float32. Finally, we use activation checkpointing.

\subsubsection{Pretraining protocol}
\label{sec:pretraining_protocol}

We train Prithvi WxC in two phases. The first phase uses 5\% drop path, a 50\% masking ratio and alternates ``local'' and ``global'' masking from gradient descent step to gradient descent step. Moreover, for each sample we select a random forecast lead time (among 0, 6, 12 and 24 hours ahead) as well as a random delta between inputs (-3, -6, -9, -12). With this randomization, we train the model on 64 A100 GPUs and batch size 1 for 100,000 gradient descent steps. After 2,500 steps of linear warm-up we perform cosine-annealing from $10^{-4}$ to $10^{-5}$. This results in a highly flexible model that we use for our downscaling and gravity wave parametrization experiments as well as for the zero-shot reconstruction evaluations.

To further attune the model to forecasting applications, we make a few changes: We reduce the masking ratio to 0\% and add a Swin-shift to the encoder. Also, we set drop path to 0\%. In addition, we fix both the forecast lead time and input delta to six hours so that there is no more randomization. Keeping the learning rate constant at $10^{-5}$, we tune the model with 1, 2 and 3 autoregressive steps on a varying compute footprint ranging from 16 to 48 GPUs. In this phase we also modify the training objective \eqref{eq:pretraining_objective} by using additional weights. For the vertical parameters, weights depend linearly on pressure level (in hPa). In addition, we weight H, $\omega$, T, U and V with 1 yet cloud, PL, QI, QL with 0.1. For the surface parameters, we weigh u10m and v10m with 1, SLP and t2m with 0.1 and the remaining parameters with 0.01. Essentially this follows \citep{lam2022graphcast} with the exception that we found it beneficial to swap the weights for t2m and u10m as well as v10m while suppressing all variables which are not standard in the AI-forecast emulation literature by a factor of ten. This version of the model is used for the forecast evaluation as well as the hurricane-forecasting use case.

\subsection{Zero-shot validation}

\subsubsection{Masked reconstruction}

\begin{figure}[h]
    \centering
    \includegraphics[width=0.9\textwidth]{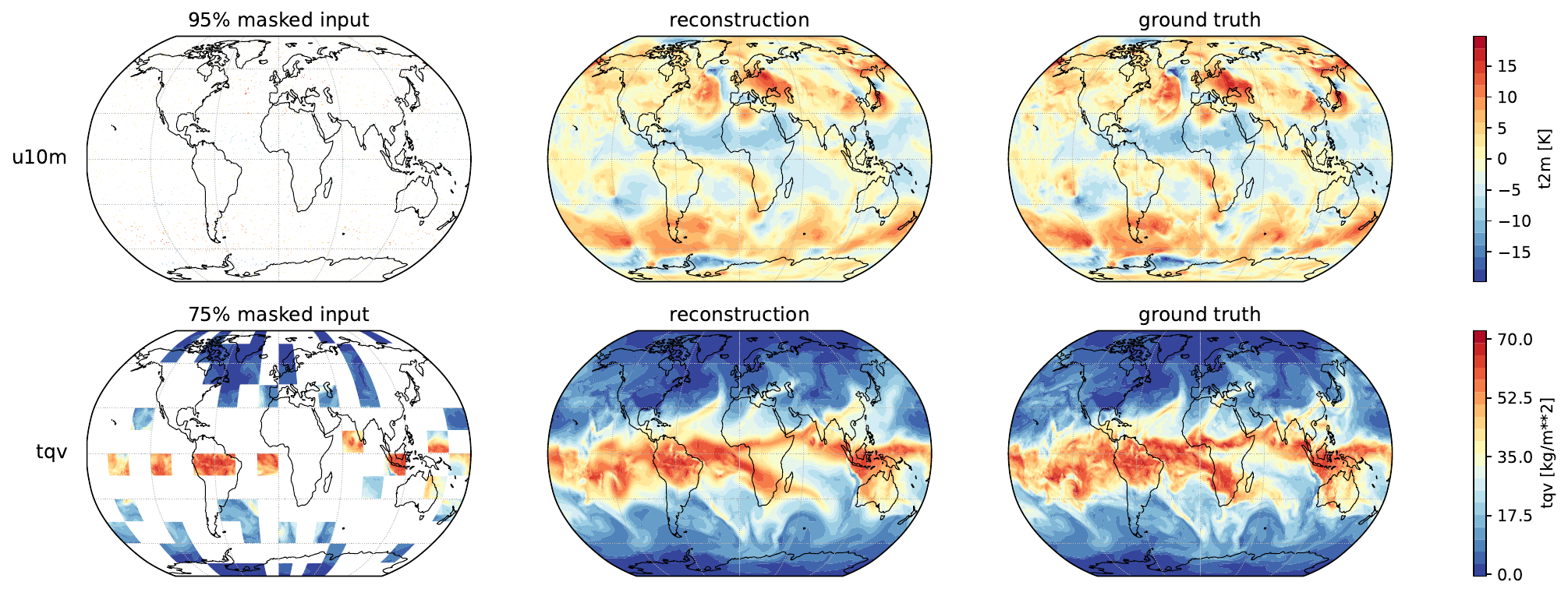}
    \caption{Zero-shot reconstruction performance with Prithvi WxC. The first row shows ``local'' masking where we mask 95\% of individual tokens. The second row shows ``global'' masking where we mask 75\% of attention windows. At these extreme masking ratios some fine structure is lost in the reconstruction. See figure \ref{fig:zero_shot_reconstruction} for metrics.}
    \label{fig:zero_shot_reconstruction_sample}
\end{figure}

\begin{figure}[h]
    \centering
    \includegraphics[width=0.9\textwidth]{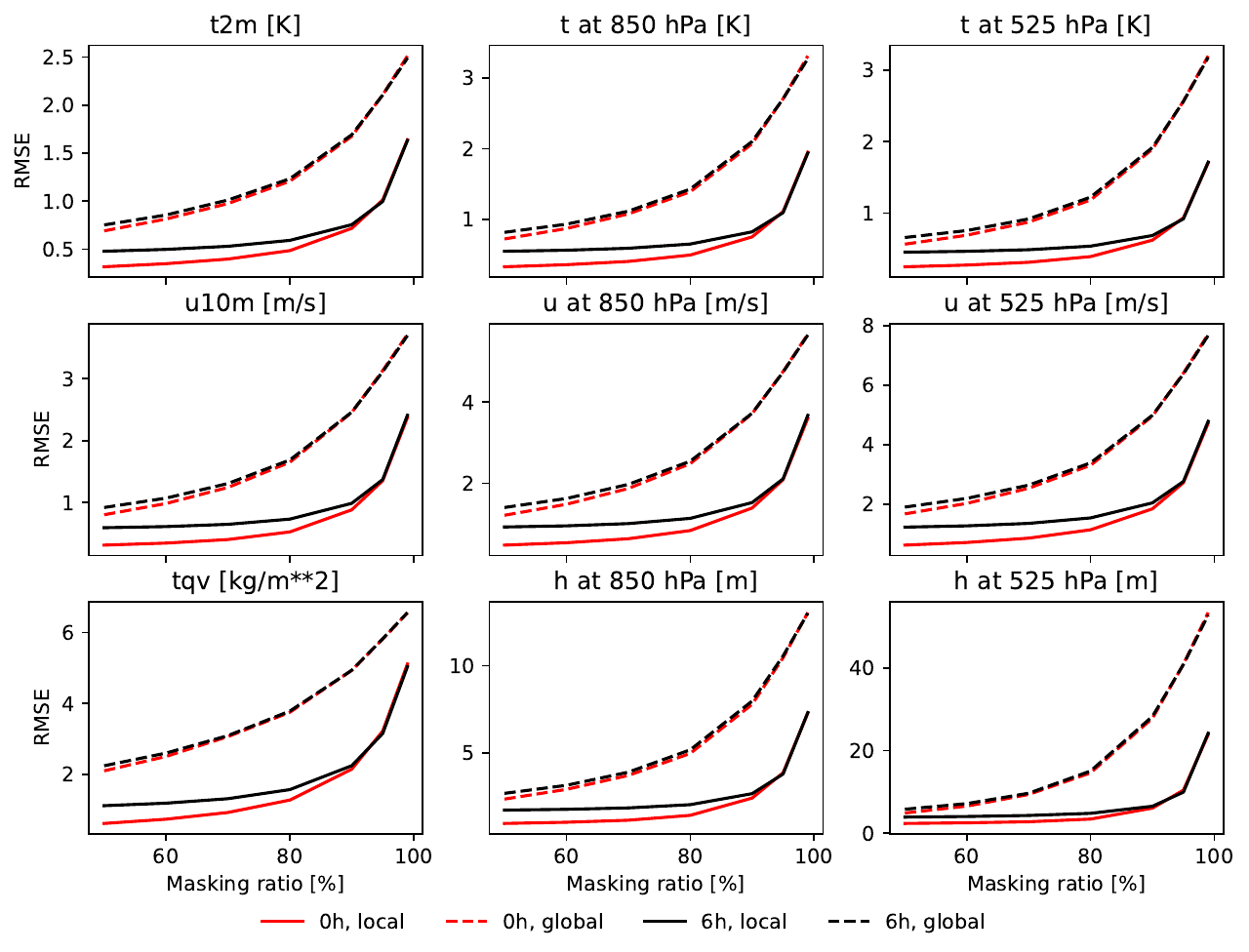}
    \caption{Zero-shot reconstruction performance of Prithvi WxC evaluated with 50, 60, 70, 80, 90, 95 and 99\% masking. Note that the 6-hour ahead values are without any forecast tuning.}
    \label{fig:zero_shot_reconstruction}
\end{figure}

To understand the model's zero shot performance, we can consider two sets of metrics: Reconstruction and forecasting. Figure \ref{fig:zero_shot_reconstruction_sample} shows two examples of the former. Note that the model is capable of reconstructing atmospheric state from as little as 5\% of the original data when the samples are still relatively dense and 25\% when we mask out large areas. Figure \ref{fig:zero_shot_reconstruction} shows RMSE scores against masking ratios for both masking strategies and 0 as well as 6 hours ahead. (The 6-hour metrics here are obtained without the additional rollout tuning phase.) It is interesting that reconstruction performance is relatively little affected by lead time at the lower end of masking ratios. This opens up the possibility of initializing a forecast model with randomly sampled tokens to obtain an ensemble forecast as well as the future research direction to fine-tune the model to integrate sparse observational data.

\subsubsection{Forecasting}

\begin{figure}[h]
    \centering
    \includegraphics[width=0.9\textwidth]{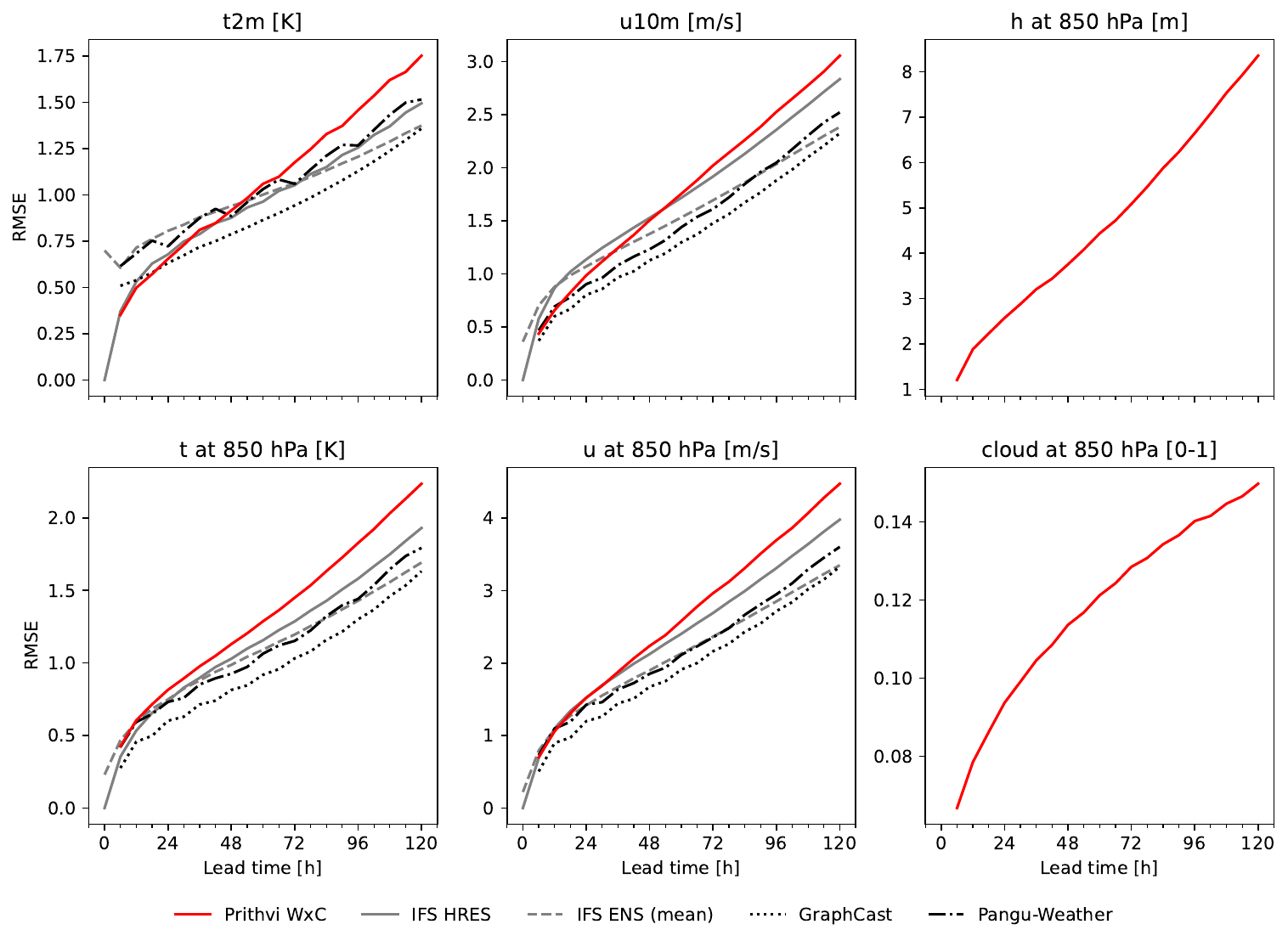}
    \caption{Zero-shot forecasting performance of Prithvi WxC.}
    \label{fig:zero_shot_forecasting}
\end{figure}

To understand the zero-shot forecasting performance of the model, we perform autoregressive forecasts with dense data up to 5 days ahead. See figure \ref{fig:zero_shot_forecasting}. To put the performance into context, we compare data from various AI forecast emulators as well as the ECMWF IFS as provided by WeatherBench2 \citep{rasp2024weatherbench}. Some care has to be taken when interpreting these results. WeatherBench2 compares against ERA5 and the IFS Analysis at 0.25 degrees resolution while we work with MERRA-2 at 0.5 by 0.625. Moreover, our model generates a number of forecasts for which no reference AI prediction exists. Most notably the ``cloud'' variables.

With all these caveats in mind, Prithvi WxC performs well to exceptionally well at very short lead times (6 and 12 hours), particularly for parameters like surface temperature. However, performance then decays and after about 66 hours Prithvi WxC falls below the performance of Pangu.

The reader might remark that we should not refer to this as zero shot performance when the model has gone through rollout tuning. However, we expect that one should do several things when truly pushing for maximal forecasting performance. Among these are adding additional convolutional or neural operator layers that improve information flow from attention window to attention window as well as deeper rollout tuning.

Finally, one might want to speculate whether the strong performance at shortest lead times is related to the masking objective, a consideration that should be of interest for nowcasting and data assimilation objectives.

\subsubsection{Hurricane track forecasting}

\begin{figure}[h]
    \centering
    \includegraphics[width=\textwidth]{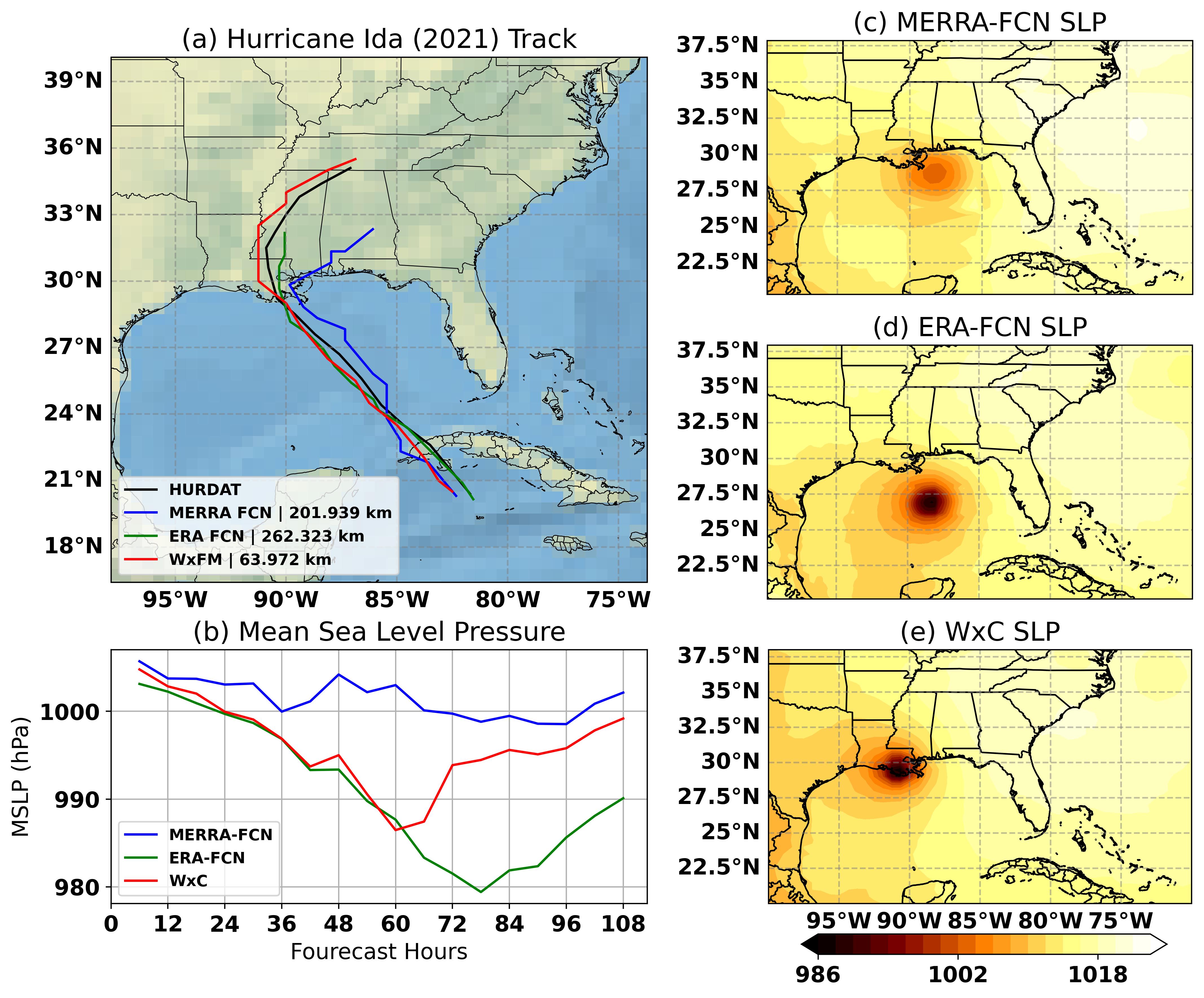}
    \caption{(a) The track of Category 4 Hurricane Ida (2021) is shown from HURDAT, MERRA-FCN (FourCastNet model trained on MERRA-2 dataset), ERA-FCN (FourCastNet model trained on ERA5 dataset), and WxC models. All models were initialized at 00 UTC on 2021-08-27. The track errors of the models, compared to the HURDAT track, are 201.9 km for MERRA-FCN, 262.32 km for ERA-FCN, and 63.9 km for WxC, as noted in the legend. (b) A 5-day forecast of Mean Sea Level Pressure (MSLP) from MERRA-FCN, ERA-FCN, and WxC models. (c-e) Spatial distribution of Sea Level Pressure (SLP) for a 60-hour forecast (valid for 12 UTC on 2021-08-29). Among the models, the WxC model predicts the hurricane landfall most accurately in terms of both spatial location and timing, compared to the HURDAT reference.}
    \label{fig:hurricane_ian}
\end{figure}

We have validated Prithvi WxC to assess its capability in forecasting the formation, dissipation, intensification, and tracking of hurricanes ranging from Category 3 to Category 5, formed over the Atlantic Ocean between 2017 and 2023. The list of hurricanes used in the analysis is provided in Table \ref{tab:hurricanes}. The performance of the model was benchmarked against observed hurricane tracks from the HURDAT database and two other models: FourCastNet trained on MERRA-2, and FourCastNet trained on the ERA5 dataset. One significant example is Hurricane Ida, a Category 4 storm that struck Louisiana in 2021. This hurricane, the second-most damaging in Louisiana’s history after Hurricane Katrina, is presented as a sample track and intensity in Figure \ref{fig:hurricane_ian}.  Prithvi WxC demonstrated superior accuracy in both track and intensity predictions. The mean track error for Prithvi WxC was 63.9 km compared to the observed tracks, significantly outperforming the MERRA-2 trained FourCastNet (201.939 km) and the ERA5 trained FourCastNet (262.323 km). Moreover, the Prithvi WxC accurately forecasted both the time and location of Ida’s landfall, with a landfall location error of less than 5 km, in contrast to errors greater than 20 km for the other models. Intensity predictions, measured in MSLP and 10-meter sustained wind speed, also favored Prithvi WxC, which outperformed the MERRA-2 trained FourCastNet and showed reasonable consistency with the ERA5 trained FourCastNet. Spatial distribution of Sea Level Pressure (SLP) for a 60-hour forecast (valid for 12 UTC on 2021-08-29) are shown the figure \ref{fig:hurricane_ian} c-e. Among the models, the WxC model predicts the hurricane landfall most accurately in terms of both spatial location and timing, compared to the HURDAT reference.

\begin{figure}[h]
    \centering
    \includegraphics[width=\textwidth]{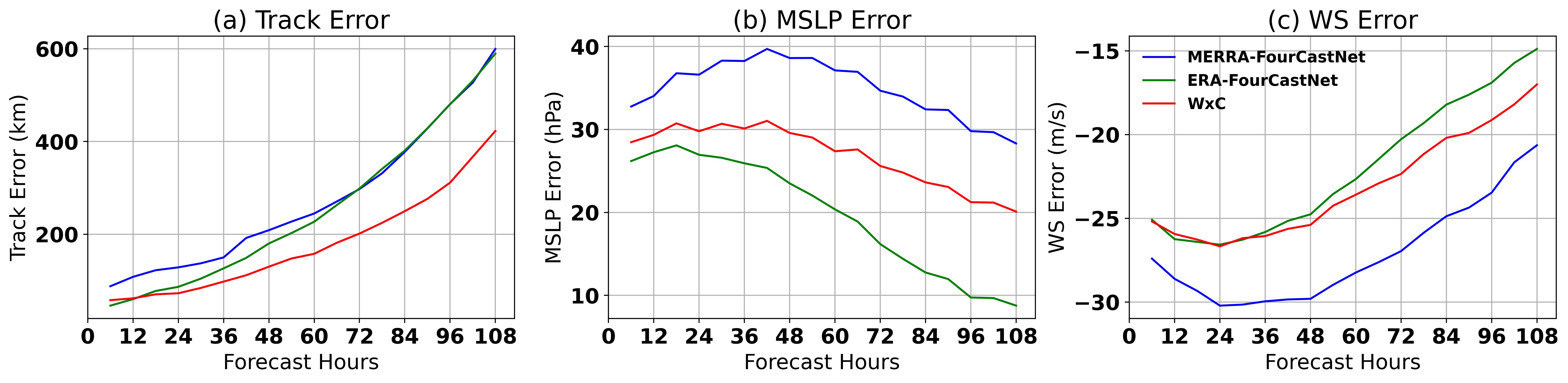}
    \caption{5 days composite (75 difference initial conditions) forecast of track errors, MSLP errors and WS errors from MERRA-FourCastNet, ERA-FourCastNet and WxC models}
    \label{fig:hurricane_composite}
\end{figure}

For a comprehensive assessment, Figure \ref{fig:hurricane_composite} presents the mean track, MSLP, and windspeed errors over a five-day forecast for all the hurricanes included in the robust evaluation (table \ref{tab:hurricanes}). Prithvi WxC consistently demonstrated lower track errors compared to both versions of FourCastNet, with the error gap increasing with longer lead times. By the end of the five-day forecast, the WxC model's track error was 200 km less than that of the FourCastNet models. While the WxC model outperformed the MERRA-2 trained FourCastNet in MSLP and windspeed predictions, it was marginally outperformed by the ERA5 trained FourCastNet, likely due to the finer spatial resolution in the ERA5 dataset. 

\section{Prithvi WxC: Downstream Validation}
\label{sec:downstream_validation}

In what follows we will look at a number of downscaling applications realized via fine-tuning. As we will see, this is frequently non-trivial. We will see changes of the dataset from MERRA-2 to ERA5 or CORDEX, changes in spatial and temporal resolution of the data, changes in selected variables and pressure levels and finally a change of spatial domain. With so much variability using the model as is and simply tuning a new head for each problem will lead to subpar results. Instead, we always add new embedding and output layers and sometimes select other architecture elements. The general pattern is that Prithvi WxC forms the typically frozen core of a model with a few additional layers that are then trained from scratch.

\subsection{Downscaling}
\label{sec:downscaling}

Downscaling models are used to refine low-resolution data to provide localized information. Several studies \citep{doury2023regional} \citep{lessig2023atmorep} \citep{nguyen2023climax} \citep{stengel2020adversarial} employ AI models as downscaling emulators to learn the relationship between low-resolution input data and high-resolution output fields. We use a pretrainedPrithvi WxC to recover the spatial structure of coarsened near surface temperature for two different datasets - MERRA-2, and CORDEX-CMIP5-RCP8.5 - with different input variables and different input resolutions.

\subsubsection{Downscaling architecture}

\begin{figure}[h]
    \centering
        \includegraphics[width=0.9\textwidth]{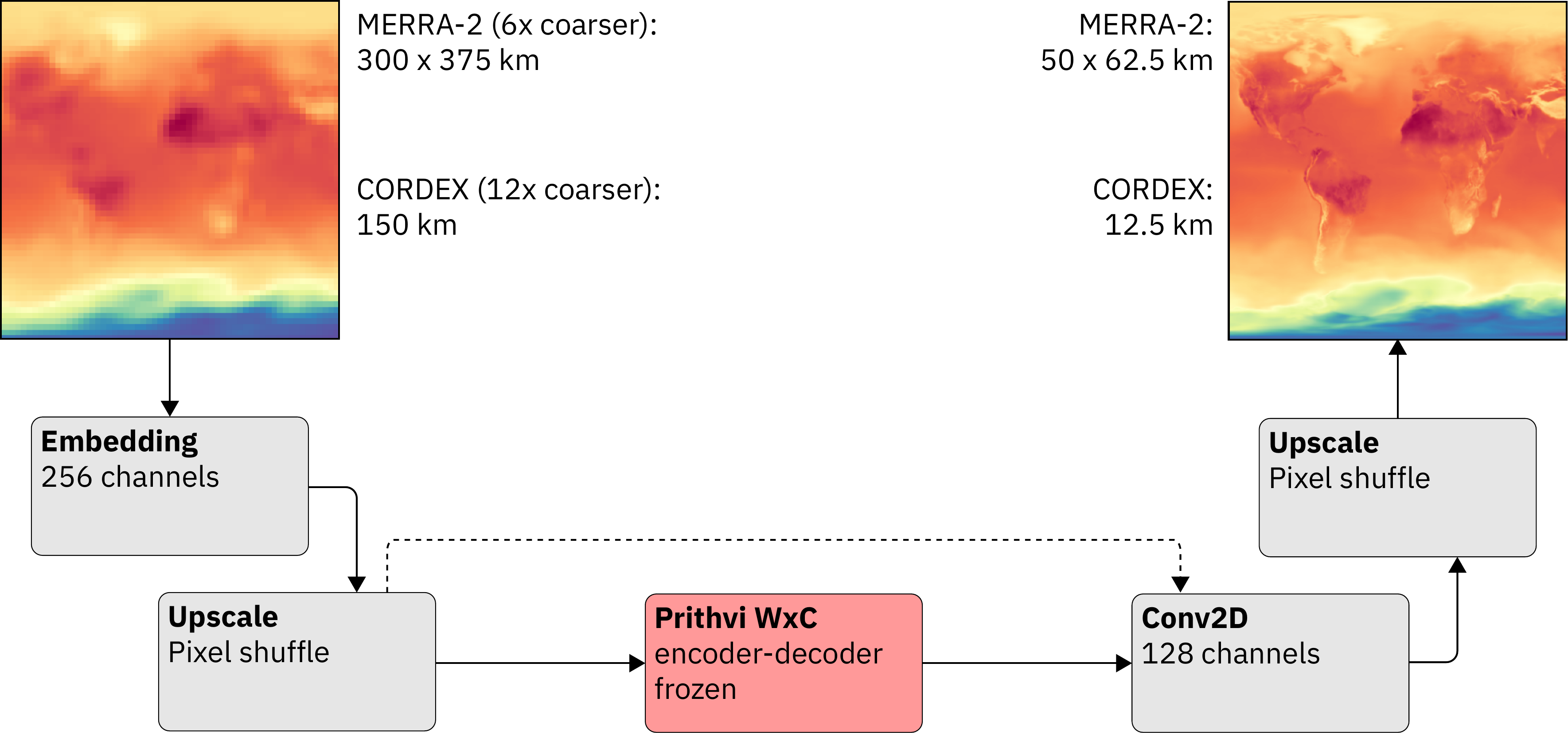}
    \caption{Finetuning Architecture of Prithvi WxC for downscaling}
    \label{fig:downscaling_arch}
\end{figure}

We use the architecture \ref{fig:downscaling_arch} to fine-tune Prithvi WxC for the downscaling task. The patch embedding layer encodes static and dynamic data for surface variables and variables at different pressure levels and optionally for multiple time steps. The first upscaling module is used for shallow feature extraction for lower frequency components and also used to control the token resolution that is input to the PrithviWxC model. This follows a deeper feature extraction by the pretrainedtransformer model. Since we set the masking ratio in the encoder to 0\,\% and the data becomes dense, we may introduce a Swin-shift in the encoder. Note that we can make this change \emph{while keeping the core transformer layers frozen}. Following \citep{liang2021swinir}, we use a convolution layer after the transformer to enhance translational equivariance, which is important in downscaling when using different local grids. The residual connection between the shallow and deep feature extraction layer allows combining lower spatial frequency information with the higher spatial frequency information. The final upscale layer focuses on extracting and refining specified output fields.

\begin{table}[]
\centering
\caption{Performance evaluation of the Prithvi WxC downscaling model compared to baselines of nearest neighbor and bilinear interpolation. Spatial RMSE (K), temporal RMSE (K), and temporal correlation is evaluated on MERRA-2 2m air temperature (T2m) for a 1-year period from 2021-01-01 to 2021-30-12 (2,912 samples); and on CORDEX near-surface air temperature (tas) for a 5-year period from 2096-01-01 to 2100-12-29 (1,829 samples) on the RCP4.5 scenario.}
\begin{tabular}{@{}lccc|ccc@{}}
\toprule
         & \multicolumn{3}{c|}{MERRA2 - T2m (K)} & \multicolumn{3}{c}{CORDEX - tas (K)} \\ \midrule
            & sp. RMSE & tp. RMSE & tp. corr.   & sp. RMSE & tp. RMSE & tp. corr.   \\
Nearest  & 3.22      & 2.46      & 0.89      & 1.89      & 1.14      & 0.99     \\
Bilinear & 3.08      & 2.34      & 0.90      & 1.47      & 0.90      & \textbf{1.00}     \\
Prithvi WxC & \textbf{0.73}    & \textbf{0.64}  & \textbf{0.98} & \textbf{0.44}    & \textbf{0.37}  & \textbf{1.00} \\ \bottomrule
\end{tabular}
\label{tab:downscaling_avg_metrics}
\end{table}

\subsubsection{Finetuning Prithvi WxC for MERRA-2 downscaling}
\label{sec:downscaling_merra2}
\begin{figure}[h]
    \centering
    \includegraphics[width=0.9\textwidth]{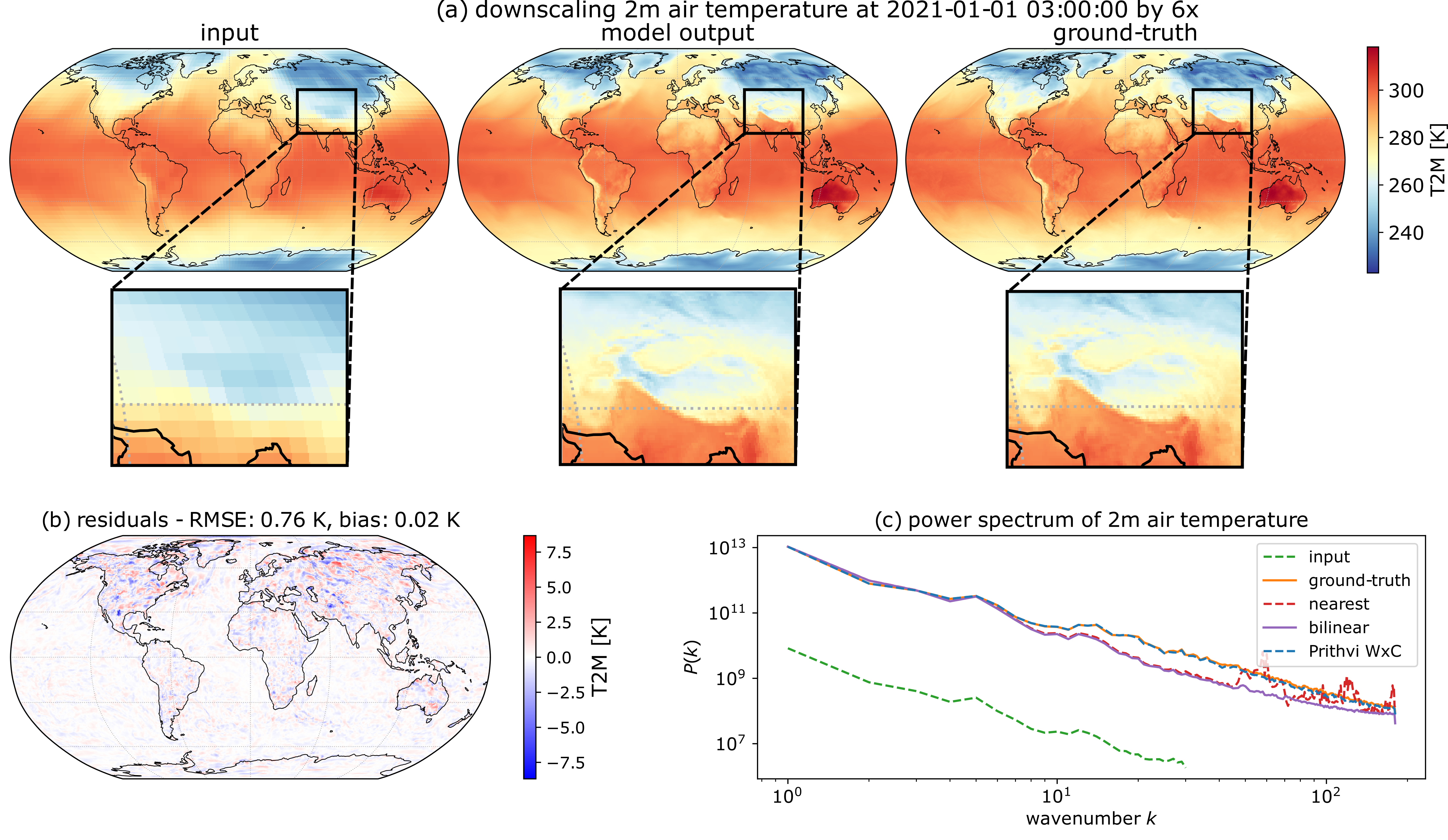}
    \caption{Downscaling MERRA-2 2m air temperature (t2m) for a sample timestamp on 2021-01-01 at 3\,UTC. (a) Visual comparison of input variable t2m, with 6\,x coarsening and smoothing; the output of the fine-tuned downscaling model; and the corresponding ground-truth. (b) Residuals between model prediction and ground-truth with RMSE of 0.76\,K. A bias of 0.02\,K indicates a negligible model overestimation. (c) Power spectra of t2m input, ground-truth, nearest-neighbor interpolation, bilinear interpolation, and downscaling model. Especially towards higher frequencies, the ground-truth power spectrum is best fit by the Prithvi WxC downscaling model.}
    \label{fig:downscaling_merra_t2m}
\end{figure}
To validate the overall downscaling performance in a clean setup that isolates model performance from dataset questions, we finetune a 6x weather downscaling model for 2m surface temperature using MERRA-2 data. The input data variables are the same as used for pre-training. We first coarsen MERRA-2 data from dimension 361 x 576 (50km x 62.5km resolution) to dimension 60 x 96 (300km x 375km resolution), and secondly apply a smoothing operation in form of a convolution with a 3x3 pixels kernel. Upscaling by a factor 2 before Prithvi WxC we increase the data resolution to 120 x 192 (150km x 187.5km). By using a patch size of 1 for tokenization, we make the token resolution similar to the token resolution that Prithvi WxC model was pretrainedon (100km x 125km). We then upscale by a factor of 3 to restore the low-resolution data to the original resolution of the 360 x 576 (50km x 62.5km).

Figure \ref{fig:downscaling_merra_t2m} visualizes the downscaling performance for a single timestamp. Following the example of the ClimateLearn benchmark \citep{nguyen2024climatelearn}, we compare the model performance with nearest neighbor and bilinear interpolation baselines. As the power spectra in Figure \ref{fig:downscaling_merra_t2m} (c) show, the interpolation baselines are poor at reconstructing the higher frequency wavenumbers of the ground truth, while the fine-tuned Prithvi WxC downscaling model is able to do so. The model performance is further evaluated on the entire validation period between 2021-01-01 and 2021-12-30 and results are summarized in Table \ref{tab:downscaling_avg_metrics}. Compared to the interpolation baselines, Prithvi WxC improves spatial and temporal RMSE values by over a factor of 4 and also shows the best temporal correlation. As the ClimateLearn benchmark reports improvements by over a factor of 2 when downscaling 2x coarsened ERA5 2m temperature, we validate with our 6x MERRA-2 downscaling experiment that we are able to finetune a high-performing Prithvi WxC downscaling model within a clean setup where all input variables are the same as in pre-training.

\subsubsection{Finetuning Prithvi WxC for CORDEX downscaling}
\label{sec:downscaling_cordex}
\begin{figure}[h]
    \centering
    \includegraphics[width=0.9\textwidth]{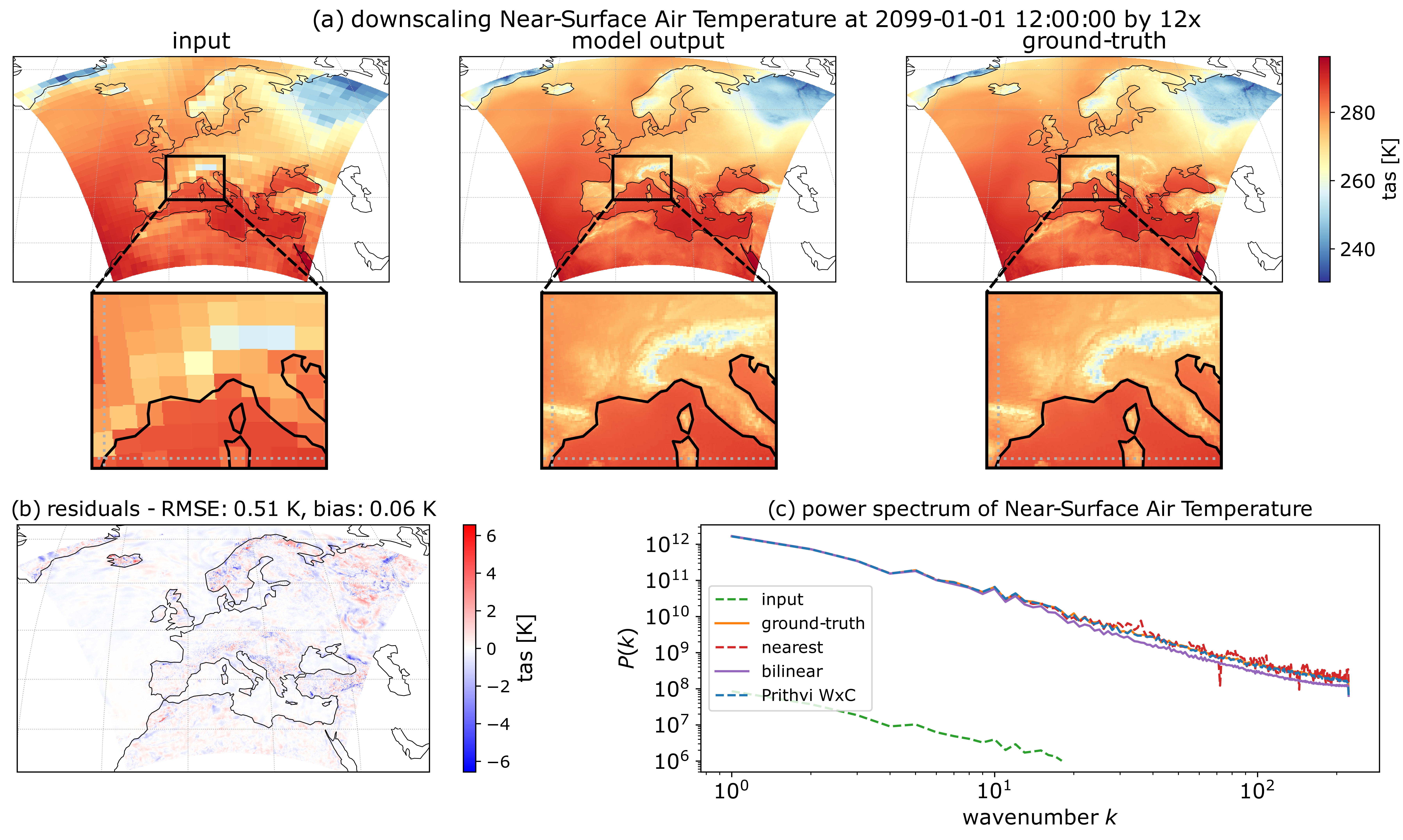}
    \caption{Downscaling CORDEX near-surface air temperature (tas) for a sample timestamp on 2099-01-01 at 12\,pm. (a) Visual comparison of input variable tas, with 12\,x coarsening and smoothing; the output of the fine-tuned downscaling model; and the corresponding ground-truth. (b) Residuals between model prediction and ground-truth with RMSE of 0.51\,K. A bias 0.06\,K indicates a slight model overestimation. (c) Power spectra of tas input, ground-truth, nearest neighbor interpolation, bilinear interpolation, and downscaling model. Especially towards higher frequencies, the ground-truth power spectrum is best fit by the Prithvi WxC downscaling model.}
    \label{fig:downscaling_cordex_t2m}
\end{figure}
We now switch from a global to a regional context as we focus on data from the Coordinated Regional Climate Downscaling Experiment (CORDEX). Specifically, we use a subset of data from the EURO-CORDEX simulations \citep{jacob_euro-cordex_2014} at a resolution of 0.11$^{\circ}$ x 0.11$^{\circ}$ (12.5\,km x 12.5\,km) covering a domain over Europe (EUR-11 CORDEX) and based on the regional climate model CNRM-ALADIN63 \citep{nabat2020modulation}, which is driven by the global climate model CNRM-CM5 \citep{voldoire2013cnrm}. In contrast to the case of previous section \ref{sec:downscaling_merra2}, this changes the dataset, the temporal step as well as the domain from the pretraining case. i.e. we are now tuning the model for a local context on a dataset that comprises significantly less inputs than the original MERRA-2 dataset. 

We finetune a 12x climate downscaling model for daily mean near-surface air temperature for a period from 2006 to 2100 under scenario RCP8.5 \citep{moss2010next}. All CORDEX input data variables (daily means) are shown in Table \ref{tab:cordex-variables}. Following \citep{doury2023regional}, we use a perfect model framework, downscaling coarsened regional climate simulations, rather than training a model to map GCM simulations to RCM simulations which may not be very well correlated. First, we coarsen the input data of dimension 444 x 444 (12.5\,km x 12.5\,km resolution) to dimension 37 x 37 (150\,km x 150\,km resolution) and apply a smoothing convolution such as for MERRA-2. One upscaling layer before the Prithvi WxC backbone increases the data resolution by a factor of 3 to a dimension of 111 x 111 (50\,km x 50\,km resolution) and two upscaling layers after the Prithvi WxC backbone increase resolution by a factor of 2 each to restore the CORDEX data's original 12.5\,km x 12.5\,km resolution. 

Model performance is evaluated on data from simulation scenario RCP4.5 which was not seen during training. Results of a single timestamp are shown in Figure \ref{fig:downscaling_cordex_t2m}. Similar as for the MERRA-2 downscaling, the power spectra in Figure \ref{fig:downscaling_cordex_t2m} (c) demonstrate better reconstruction of higher frequency wavenumbers by the finetuned Prithvi WxC downscaling model compared to the interpolation baselines. We evaluated the model performance over a time horizon of 5 years from 2096-01-01 to 2100-12-29. The average metrics displayed in Table \ref{tab:downscaling_avg_metrics} indicate improvements of spatial and temporal RMSE values by factors of around 3. Temporal correlation values are generally high across all methods which is most likely explained by the fact that we are downscaling daily mean values of near-surface air temperature. In their \textit{perfect model world} experiment, \cite{doury2023regional} report a mean spatial RMSE of 0.55\,K when including the target variable (tas) as input feature. Our mean spatial RMSE is 0.44\,K, calculated on a bigger spatial domain (corresponding approximately to the EURO-CORDEX simulation domain) without masking the sea and over a shorter time period of 5-years. Bearing these differences in mind, we are confident that finetuning the Prithvi WxC downscaling model with \textit{frozen} MERRA-2 pretrained backbone on a new dataset of distinct nature resulted in a competitive downscaling model. Future work will encompass downscaling from a global climate model to a regional climate model, where the true advantage of AI-based climate emulators can come to light.

\subsection{Climate Model Parameterization for Gravity Wave Flux}

This task uses the pretrained Prithvi WxC to create a fine-tuned model for climate applications. The fundamental question being: can we (re-)use large AI models to develop improved, data-driven climate model parameterizations for small-scale atmosphere-ocean processes?

\textbf{Background:} Atmospheric gravity waves (GWs) are intermittent, small-scale ($\mathcal{O}$(1) to $\mathcal{O}$(1000) km) perturbations generated around thunderstorms, jet disturbances, flow over mountains, etc. \citep{Fritts.Alexander2003, Achatz.etal2023}. Gravity waves couple the different layers of the atmosphere by carrying surface momentum to stratospheric and even mesospheric heights. Yet, most climate models fail to resolve them owing to limited resolution. Thus, they belong to a class of key physical processes crucial to the earth's momentum budget but only crudely represented in coarse-climate models using inadequate \emph{physical parameterizations}. %Their approximate single-column representation  called a physical parameterizationoften ignores one or more key physical properties: principles like horizontal propagation, transient evolution, source/sink spectrum, etc. \citep{Plougonven.etal2020}.

An improved parametric representation of gravity waves in comprehensive climate models can potentially improve the representation of the seasonal transitions \citep{McLandress.etal2012}, clear air turbulence \citep{Plougonven.Zhang2014}, Antarctic extreme heat \citep{Choi.etal2024}, and tropical predictability \citep{Baldwin.etal2001}; leading to more certain climate predictions and advancements in mechanistic understanding.

From an AI perspective, this downstream prediction task moves from predicting the large-scale atmospheric state prediction to smaller-scale state prediction, and leverages the cross-scale learning from pre-training. As such, the finetuning task is defined to use the latent space of Prithvi to develop data-driven physical parameterizations to provide missing sub-grid scale variability in coarse-climate models at zero-lag. This is somewhat akin to the downscaling task where CORDEX is used to augment missing sub-grid information during fine-tuning. For this task, the model is fine-tuned using high-fidelity, high-resolution gravity wave data extracted from ERA5 (which resolves a substantial portion of the atmospheric gravity waves, if not all).

% \textbf{Connection to model parameterizations in general.}
% Climate model parameterizations are the leading source of uncertainty in accurate climate prediction, yet the development of new, more accurate physical parameterizations has been challenging, either due to limited observations to validate them, or due to computational challenges associated with coupling a complex reduced-order model into a sophisticated climate model.

% Parameterization development can leverage modern advances in AI to (a) develop improved, data-driven parameterizations by training on high-fidelity data that accurately captures the process physics, or (b) improve climate model speed by reducing latency due to traditional parameterization.

% Since GWs are a purely dynamical phenomena common to simple to intermediate to comprehensive climate models alike, we choose them as a test case to define a climate model parameterization focused fine-tuning task. This way, the fine-tuning task can leverage the learning from both the transformer backbone and the high-quality high-resolution fine tuning data.

\subsubsection{Extracting GW data for finetuning.}

The goal is to accurately predict the momentum fluxes carried by waves generated in different parts of the globe by different processes, given the background atmospheric  state. The approach is similar to that followed by traditional single-column parameterizations \citep{Lott.Miller1997, Scinocca2003, Kim.etal2003}. Here, we do so by learning from high-resolution data. In very simple terms, given the background atmospheric state around a mountain (e.g., Andes), or around tropical storm, can our ML model predict whether the waves are spontaneously generated, and if they are, calculate the net momentum fluxes they carry; not unlike predicting the cloud cover for a given set of atmospheric conditions.

We use four years of ERA5 global reanalysis on 137 model vertical levels and 30 km horizontal resolution at hourly-frequency to prepare the training data for fine-tuning. The top 15 levels, i.e., levels above 45 km are removed due to artifical sponge damping in effect, so effectively 122 vertical levels. The model takes the zonal wind speed ($u$), meridional wind speed ($v$), the temperature ($T$), and pressure ($p$), along with positional variables latitude, longitude, and surface height as input. These variables collectively describe the background state of the atmosphere. The model outputs the directional momentum fluxes carried by gravity waves. These fluxes describe the net instantaneous momentum the gravity waves carry. These directional fluxes are mathematically expressed as the covariances $(u'\omega',v'\omega')$, and are computed using Helmholtz decomposition using the horizontal ($u$,$v$)=(U,V) and vertical wind speeds ($\omega$=OMEGA). Both the input and output are conservatively coarse-grained to a 64$\times$128 ($\approx$ 300 km) latitude-longitude grid to be consistent with a typical coarse-climate model and to remove phase dependencies of the calculated fluxes.
\begin{figure}[!htb]
    \centering
    \includegraphics[width=0.9\textwidth]{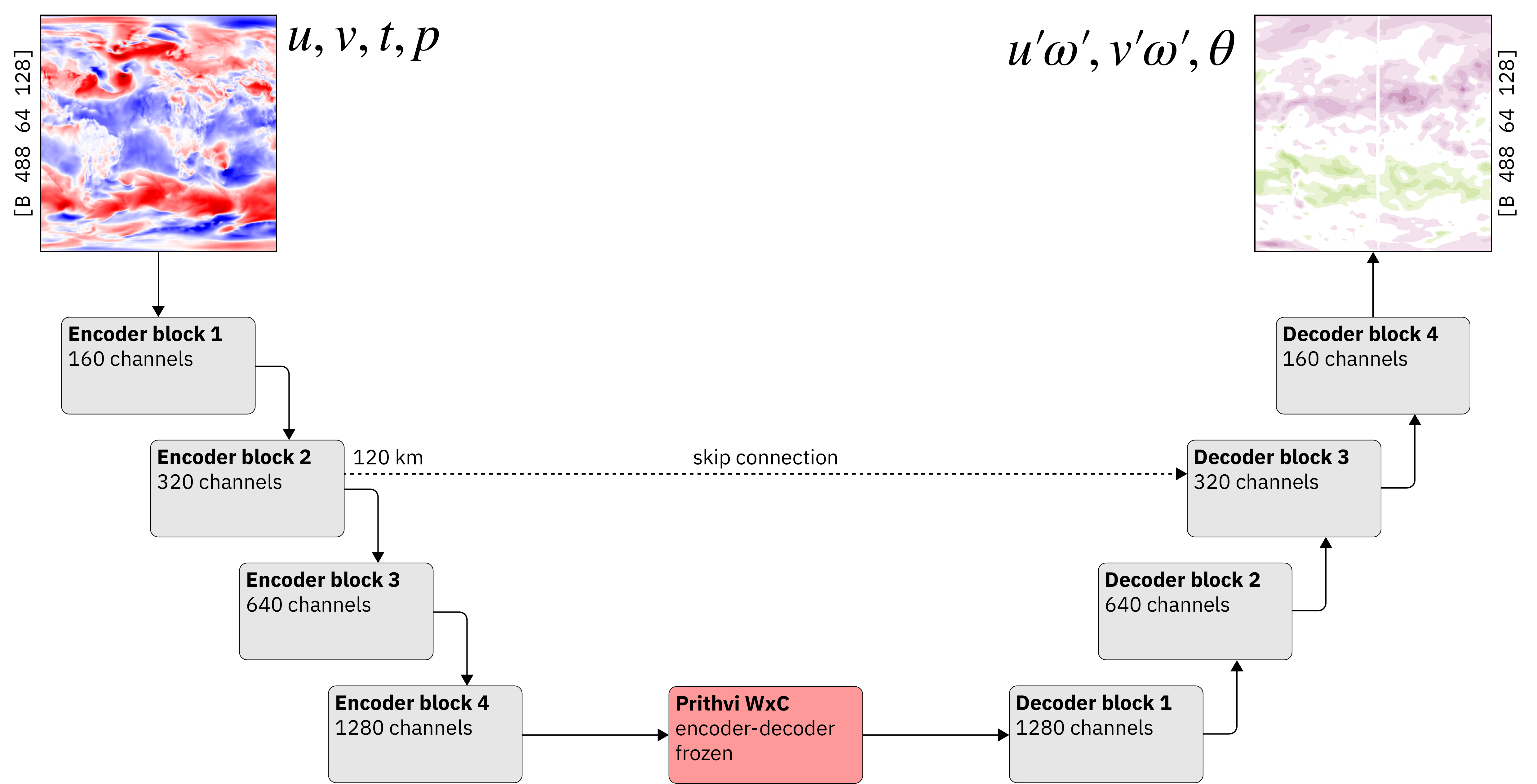}
    \caption{Finetuning Architecture of Prithvi WxC for parameterization of gravity wave flux}
    \label{fig:gravitywave}
\end{figure}

\subsubsection{Finetuning Prithvi WxC}

%An Attention U-Net model was trained on the same dataset to create an advanced baseline compared to standard MLP. The approach has also been applied in previous studies  \citep{Oktay.etal2018, TREBING2021178, gupta2024machine}.

The architecture schematic for the finetuning is shown in Figure \ref{fig:gravitywave}. During fine-tuning Prithvi WxC, we freeze the encoder and decoder part of the model. The frozen encoder is preceded by 4 learnable convolution blocks each with an increasing number of hidden channels, i.e., $C$, 2$C$, 4$C$ and then 8$C$, where $C$ = 160. Likewise, the frozen decoder is succeeded by 4 new learnable convolution blocks. Since gravity wave flux prediction is an instantaneous flux calculation task, we fix the lead time $\delta t$ to zero. The instantanous model input for fine-tuning has shape [1, 488, 64, 128] where the 488 channels comprise the four background variables $u$, $v$, $t$ and $p$ on 122 vertical levels each, and on a 64 $\times$ 128 horizontal grid, as discussed above. The model was fine-tuned to produce an output with shape [366, 64, 128] comprising of the potential temperature, $u'\omega'$, and $v'\omega'$ on 122 vertical levels each.

The fine-tuning model leveraged a U-Net like architecture to allow the model to extract high-frequency information from the given data source. We re-emphasize that Prithvi WxC was pretrained on the MERRA-2 dataset but for fine-tuning we are using the downscaled ERA5 dataset. More importantly, the finetuned model uses global information as input to predict global fluxes as output, providing a direct contrast to traditional single-column parameterizations. 
Access to global information allows the model to learn the horizontal propagation of gravity waves.% - a vital feature shared by most parameterized physical processes but completely ignored by current parameterizations.

\subsubsection{Results and description}

\begin{figure}[!htb]
    \centering
    \includegraphics[width=1\textwidth]{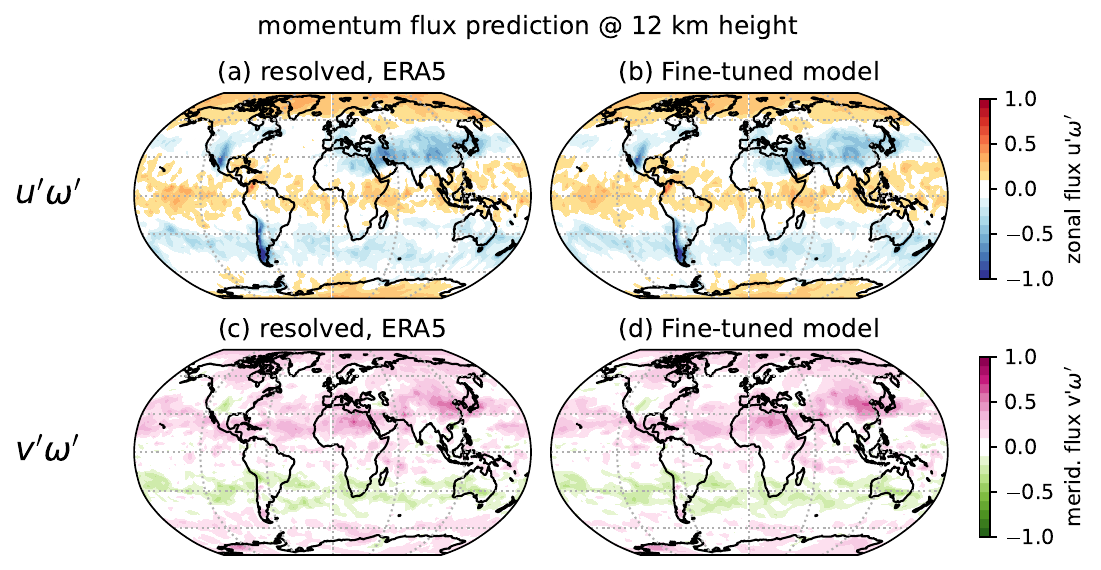}
    \caption{True vs. predicted (non-dimensionalized) momentum fluxes in the upper troposphere (12 km height) for the gravity wave flux parameterization downstream task. (a) and (c) respectively show the fluxes$u'\omega'$ and $v'\omega'$ respectively from ERA5, and (b) and (d) show the respective predictions from the fine-tuned model. All fluxes are monthly averaged for May 2015. The vertical derivative of the fluxes represents the wind-forcing tendencies due to gravity waves in the atmosphere and can be used to represent a portion of unresolved sub-grid tendencies in climate models.}
    \label{fig:results_gravity_wave}
\end{figure}
%Similar to how a precipitation parameterization would be expected to accurately predicted when it would rain and how much it would rain, a good data-driven GW parameterization would correctly identify where GWs are generated and how much momentum they carry.

As a straightforward test, we look at the climatological distribution, i.e., the monthly-averaged momentum fluxes in the upper troposphere, and compare the spatial distribution of the predicted directional fluxes with the validation data from ERA5 (Figure \ref{fig:results_gravity_wave}). The prediction from the baseline closely agrees with the true flux distribution in the upper troposphere. The nature and properties of the waves over land can be significantly different from waves over the ocean. Therefore, getting a strong agreement over both the ocean and the land indicates effective learning. For instance, enhanced fluxes over the Rocky Mountains, the Andes, and the Himalayas indicates the finetuned model skillfully predicts the stationary waves generated over mountain ranges. Likewise, the tropical band of positive flux (in Figure \ref{fig:results_gravity_wave}b) in the tropics points to effective learning of non-stationary gravity waves generated around intense convective and precipitation systems. In fact, the finetuning model even outperforms our task-specific baselines created using MLPs and Attention U-Nets.

\subsubsection{Scientific importance and broader impact}
Without loss of generality, the same finetuning procedure can be applied to develop parameterizations for other sub-grid atmospheric processes of relevance to climate; albeit with some tweaks. A coarse climate model with a typical resolution of $\mathcal{O}$(100) km fails to capture most gravity wave effects (or clouds, or fine-scale turbulence) due to its inability to resolve the smaller-scales. Owing to periodic data assimilation and higher-resolution, numerical weather prediction models are largely unaffected by these biases. Running climate models at a high-resolution over multiple centuries, however, is computationally not so feasible. To address this, we have proposed one climate-focused application of Prithvi WxC and demonstrated its effectiveness. This model can subsequently be integrated with coarse-resolution climate models of varying complexity to account for the ``missing" gravity wave physics and correct the physics tendencies. The accuracy of the predicted fluxes also points to the remarkable effectiveness of the fine-tuning process in blending task-specific data from heterogenous sources.

\section{Discussion and conclusions}

The development of accurate and efficient weather and climate models is crucial for understanding and predicting Earth's complex atmospheric-oceanic system. While traditional numerical weather prediction (NWP) models have made significant strides, they require substantial computational resources. The emergence of deep learning, particularly foundation models pretrained on vast datasets, offers a promising alternative for weather and climate modeling.

This study introduces Prithvi WxC, a 2.3 billion parameter foundation model designed for weather and climate applications. Trained on 160 atmospheric variables from NASA's Modern-Era Retrospective Analysis for Research and Applications, Version 2 (MERRA-2) dataset, Prithvi WxC leverages a scalable and flexible transformer-based architecture to capture both regional and global dependencies in atmospheric data. Unlike task-specific deep learning models, Prithvi WxC aims to address a diverse set of downstream tasks, aligning with the foundation model paradigm prevalent in AI research. To achieve this, the model introduces a new architecture and novel objective function. The latter combines masked reconstruction with forecasting, incorporating climatological information to enhance its generalizability.

The zero-shot evaluation introduces reconstruction as a new benchmark, revealing that the model excels in forecasting at shorter lead times. We hypothesize that this strength stems from the masking objective, which encourages Prithvi WxC to grasp atmospheric dynamics with limited temporal progression.

When it comes to fine-tuning, it's important to highlight the diversity of datasets, parameters, and resolutions addressed in the downscaling and parameterization examples. In both cases, we demonstrate that a pretrained, frozen transformer trained on a single dataset can be effectively combined with additional architectural components to achieve strong results on new tasks with different datasets. Furthermore, the CORDEX downscaling case showcases the model's ability to operate in both global and regional contexts, a characteristic that we attribute to the heavy use of ``global'' masking during pretraining.

Even though there is no previous work on AI-based downscaling using MERRA-2, we chose this example to isolate the model's and architecture's downscaling performance from questions of distribution shift when changing datasets. Here, we found that the fine-tuned Prithvi WxC model improves by more than a factor of 4 over interpolation baselines. This 6x downscaling compares to an improvement factor of 2 when doing 2x downscaling with ERA5 data in the ClimateLearn benchmarks. That is, we have doubled the performance for a threefold resolution increase, evidence of strong performance. This is mirrored by the more applicable CORDEX example which compares favorably to the results of \cite{doury2023regional}.

Finetuning Prithvi WxC also demonstrates that large transformer-based foundation models can effectively learn mesocale atmospheric evolution, helping to streamline, enhance, and accelerate the development of physical parameterizations in climate models, which in turn improves prediction accuracy on interannual timescales. The fine-tuned model produces significantly improved predictions across all six hotspots, including both the relatively smoother fluxes over the Andes, Southern Ocean, Newfoundland, and the Scandinavian Mountains, as well as the more turbulent fluxes over the Pacific Ocean and Southeast Asia. Notably, for the Andes (mountain waves) and the Southern Ocean (non-mountain waves), the fine-tuned model achieves correlation coefficients of 0.99 and 0.97, respectively, when compared to the observed fluxes.

The latent encoder-decoder space of Prithvi WxC foundation model captures a comprehensive understanding of atmospheric evolution by training on vast amounts of data, including winds, temperature, humidity, radiation, and soil moisture. Instead of building task-specific ML-models from scratch, these pretrained encoders can be used to develop more precise data-driven models of atmospheric processes.

\subsubsection*{Acknowledgments}

We would like to thank S.~Karthik Mukkavilli who contributed in the early stages of this project. We would also like to thank Shubha Ranjan from NASA Advanced Supercomputing (NAS) Division, and Mike Little from Goddard Spaceflight Center for their help and support, and McKenzie Hicks and Elizabeth Fancher from NASA IMPACT as well as Kathy Duviella from IBM Research for project management support. Finally, we would like to thank Wei Ji Leong and Raghu Kiran Ganti.

V. Anantharaj is supported by the Office of Science of the U.S. Department of Energy under Contract No. DE-AC05-00OR22725. Aditi Sheshadri and Aman Gupta are supported by Schmidt Sciences, LLC, a philanthropic initiative founded by Eric and Wendy Schmidt, as part of the Virtual Earth System Research Institute (VESRI). Aditi Sheshadri also acknowledges support from the National Science Foundation through Grant OAC‐2004492.

This work was supported by NASA’s Office of Chief Science Data Officer and Earth Science Division’s Earth Science Scientific Computing, Earth Science Data Systems Program, and the Earth Science Modeling and Analysis Program.

\section*{Data Availability}

The code for the Prithvi WxC model is available at \url{https://github.com/NASA-IMPACT/Prithvi-WxC}. The fine-tuning code for climate model parameterization for gravity wave Flux is available at \url{https://github.com/NASA-IMPACT/gravity-wave-finetuning}. The model checkpoints and sample data are available at \url{https://huggingface.co/Prithvi-WxC}.

\bibliography{iclr2025_conference}
\bibliographystyle{iclr2025_conference}

\appendix

\section{Tables}

\begin{table}[h!]
\centering
\caption{List of Surface Variables}
\begin{tabular}{c|c|c}
\toprule
\textbf{Variable} & \textbf{Collection} & \textbf{Description} \\ \midrule
U10 & M2I1NXASM & 10 m zonal wind \\ \hline
V10 & M2I1NXASM & 10 m meridional wind \\ \hline
T2M & M2I1NXASM & 2 m surface temperature \\ \hline
QV2M & M2I1NXASM & 2 m specific humidity \\ \hline
PS & M2I1NXASM & Surface Pressure \\ \hline
SLP & M2I1NXASM & Sea Level Pressure \\ \hline
TS & M2I1NXASM & Skin Temperature \\ \hline
TQI & M2I1NXASM & Column-total ice  \\ \hline
TQL & M2I1NXASM & Column-total liquid water  \\ \hline
TQV & M2I1NXASM & Column-total watre vapor  \\ \hline

GWETROOT & M2T1NXLND & Rootzone soil wetness relative to soil holding capacity \\ \hline
LAI & M2T1NXLND & Leaf area index  \\ \hline
EFLUX & M2T1NXFLX & Surface latent heat flux  \\ \hline
HFLUX & M2T1NXFLX & Surface sensible heat flux  \\ \hline
Z0M & M2T1NXFLX & Surface roughness \\ \hline
LWGEM & M2T1NXRAD & Longwave radiation emitted by the surface  \\ \hline
LWGAB & M2T1NXRAD & Longwave radiation absorbed by the surface  \\ \hline
LWTUP & M2T1NXRAD & Upward longwave at the top of atmosphere  \\ \hline
SWGNT & M2T1NXRAD & Net downward shortwave radiation at the surface \\ \hline
SWTNT & M2T1NXRAD & Net shortwave at top of atmosphere  \\ \bottomrule
\end{tabular}
\label{tab:variables}
\end{table}

\begin{table}[h!]
\centering
\caption{List of Native Vertical Level Variables}
\begin{tabular}{c|c|c}
\toprule
\textbf{Variable} & \textbf{Collection} & \textbf{Description} \\ 
\midrule
U & M2I3NVASM & Wind speed/direction \\ \hline
V & M2I3NVASM & Wind speed/direction \\ \hline
OMEGA & M2I3NVASM & Vertical motions \\ \hline
T & M2I3NVASM & Air temperature \\ \hline
QV & M2I3NVASM & Specific humidity \\ \hline
PL & M2I3NVASM & Actual mid-level pressure \\ \hline
H & M2I3NVASM & Mid-layer height (equivalent to the geopotential height) \\ \hline
CLOUD & M2I3NVASM & Cloud fraction at this layer for radiation \\ \hline
QI & M2I3NVASM & Cloud mass fraction that is ice  \\ \hline
QL & M2I3NVASM & Cloud mass fraction that is water  \\ \hline
\hline
\multicolumn{3}{|c|}{\textbf{Nominal Pressure (hPa)}} \\ \hline
\multicolumn{3}{|c|}{985 \textbar{} 970 \textbar{} 925 \textbar{} 850 \textbar{} 700 \textbar{} 600 \textbar{} 525 \textbar{} 412 \textbar{} 288 \textbar{} 245 \textbar{} 208 \textbar{} 150 \textbar{} 109 \textbar{} 48} \\ \bottomrule
\end{tabular}
\label{tab:variables2}
\end{table}

\begin{table}[h!]
\centering
\caption{List of Static Variables}
\begin{tabular}{c|c|c}
\toprule
\textbf{Variable} & \textbf{Dataset} & \textbf{Description} \\ \midrule
PHIS & M2C0NXASM & Surface geopotential height \\ \hline
FRLAND & M2C0NXASM & Fraction of surface that is land \\ \hline
FROCEAN & M2CONXCTM & Fraction of surface that is ocean \\ \hline
FRACI & M2CONXCTM & Fraction of surface that is ice \\ \bottomrule
\end{tabular}
\label{tab:variables3}
\end{table}

\begin{table}[]
\centering
\caption{List of CORDEX variables. Experiments use daily mean values of scenario simulations RCP4.5 and RCP8.5 between 2006 and 2100.}
\label{tab:cordex-variables}
\begin{tabular}{|l|l|l|l|}
\toprule
\textbf{Variable}        & \textbf{Level (hPa)} & \textbf{Unit}              & \textbf{Description}         \\ \midrule
hus500, hus700,   hus850 & 500, 700, 850        & -   & Specific Humidity            \\ \hline
ta500, ta700, ta850      & 500, 700, 850        & K  & Air Temperature              \\ \hline
ua500, ua700, ua850      & 500, 700, 850        & {m/s} & Eastward Wind                \\ \hline
va500, va700, va850      & 500, 700, 850        & m/s & Northward Wind               \\ \hline
zg500, zg700, zg850      & 500, 700, 850        & m   & Geopotential Height          \\ \hline
psl                      & surface              & Pa  & Sea Level Pressure           \\ \hline
tas                      & surface              & K   & Near-Surface Air Temperature \\ \hline
uas                      & surface              & m/s & Eastward Near-Surface Wind   \\ \hline
vas                      & surface              & m/s & Northward Near-Surface Wind  \\ \bottomrule
\end{tabular}
\end{table}

\begin{table}
\centering
\caption{List of Hurricanes for Evaluation}
\label{tab:hurricanes}
\begin{tabular}{l | c | c | p{7cm}}
\toprule
     Name (YYYY) & Category &  \#IC &                                 Initial Conditions \\
\midrule
     Jose (2017) &       C4 &    4 &     2017090900, 2017091000, 2017091100, 2017091200 \\
   Harvey (2017) &       C4 &    2 &                             2017082400, 2017082500 \\
     Irma (2017) &       C5 &    3 &                 2017090500, 2017090600, 2017090700 \\
  Michael (2018) &       C5 &    2 &                             2018100800, 2018100900 \\
 Florence (2018) &       C4 &    4 &     2018091000, 2018091100, 2018091200, 2018091300 \\
   Dorian (2019) &       C5 &    4 &     2019083100, 2019090100, 2019090200, 2019090300 \\
  Lorenzo (2019) &       C5 &    4 &     2019092500, 2019092600, 2019092700, 2019092800 \\
 Humberto (2019) &       C3 &    2 &                             2019091400, 2019091500 \\
    Delta (2020) &       C4 &    2 &                             2020100600, 2020100700 \\
    Laura (2020) &       C4 &    2 &                             2020082300, 2020082400 \\
     Iota (2020) &       C4 &    2 &                             2020111400, 2020111500 \\
     Zeta (2020) &       C3 &    1 &                                         2020102500 \\
      Eta (2020) &       C4 &    2 &                             2020110700, 2020110800 \\
    Teddy (2020) &       C4 &    5 &  2020091400, 2020091500, 2020091600, 2020091700,  2020091800\\
      Ida (2021) &       C4 &    3 &                 2021082700, 2021082800, 2021082900 \\
    Grace (2021) &       C3 &    2 &                             2021081700, 2021081800 \\
    Larry (2021) &       C3 &    5 &  2021090200, 2021090300, 2021090400, 2021090500,  2021090600\\
      Sam (2021) &       C4 &    5 &  2021092500, 2021092600, 2021092700, 2021092800,  2021092900\\
      Ian (2022) &       C5 &    4 &     2022092500, 2022092600, 2022092700, 2022092800 \\
    Fiona (2022) &       C4 &    4 &     2022091600, 2022091700, 2022091800, 2022091900 \\
 Franklin (2023) &       C4 &    1 &                                         2023082200 \\
      Lee (2023) &       C5 &    8 &  2023090500, 2023090600, 2023090700, 2023090800, 2023090900, 2023091000, 2023091100, 2023091200  \\
   Idalia (2023) &       C4 &    4 &     2023082700, 2023082800, 2023082900, 2023083000 \\
\bottomrule
\end{tabular}
\end{table}

\end{document}